\documentclass[10pt,twocolumn,letterpaper]{article}
\usepackage{times}
\usepackage{graphicx} 
\usepackage{subfigure} 
\usepackage{amsmath}
\usepackage{epstopdf}
\usepackage{float}
\usepackage{courier}
\usepackage{cite}
\usepackage{multirow}
\usepackage{dblfloatfix}
 
\usepackage[utf8]{inputenc}
\usepackage[english]{babel}
 
\usepackage{algorithm}
\usepackage{algorithmic}

\usepackage{hyperref}

\usepackage{iccv}
\iccvfinalcopy
\usepackage[section]{placeins}


\begin{document}

\title{\textbf{Sepsis Prediction and Vital Signs Ranking in Intensive Care Unit Patients}}

\author{Avijit Mitra$^1$, Khalid Ashraf$^2$\\
$^1$Semion, Mohakhali DOHS, Dhaka 1206, Bangladesh.\\
$^2$Semion, Palo Alto, CA 94303, USA.\\
{\tt\small avipartho@gmail.com, khalid@semion.ai }
}

\maketitle

\begin{abstract}

We study multiple rule-based and machine learning (ML) models for sepsis detection. We report the first neural network detection and prediction results on three categories of sepsis. We have used the retrospective Medical Information Mart for Intensive Care (MIMIC)-III dataset, restricted to intensive care unit (ICU) patients. Features for prediction were created from only common vital sign measurements. We show significant improvement of AUC score using neural network based ensemble model compared to single ML and rule-based models. For the detection of sepsis, severe sepsis, and septic shock, our model achieves an AUC of 0.97, 0.96 and 0.91, respectively. Four hours before the positive hours, it predicts the same three categories with an AUC of 0.90, 0.91 and  0.90 respectively. Further, we ranked the features and found that using six vital signs consistently provides higher detection and prediction AUC for all the models tested. Our novel ensemble model achieves highest AUC in detecting and predicting sepsis, severe sepsis, and septic shock in the MIMIC-III ICU patients, and is amenable to deployment in hospital settings. 

\end{abstract}

\section{Introduction}
\label{intro}

Sepsis, a common systemic response to infection, is one of the leading causes of death in the United States \cite{murphy2013deaths}. Each year, approximately 750,000 hospitalized patients are diagnosed with severe sepsis in the United States alone and the mortality rate may go up to one-third of this population \cite{angus2001epidemiology,stevenson2014two}. The treatment cost is very high, \$20.3 billion dollars annually in USA hospitals \cite{pfuntner2006costs}. Compared to any other condition, this is, on average, almost twice the cost \cite{TheCosto23:online}. Moreover, the occurrence of severe sepsis is increasing by an approximate 13\% per year\cite{gaieski2013benchmarking}. With the progress of sepsis, organ failure and eventually death becomes evident. Previous studies have showed that through early diagnosis and treatment, it's possible to reduce mortality as well as the related medical expenditures \cite{rivers2001early,nguyen2007implementation,kumar2006duration}. 

According to the Sepsis-3 definition \cite{singer2016third}, sepsis is “life-threatening organ dysfunction caused by a dysregulated host response to infection ”. Traditionally, a person is diagnosed with sepsis if he or she demonstrates two or more Systemic Inflammatory Response Syndrome (SIRS) \cite{levy20032001} criteria with a suspected infection. This turns into severe sepsis in the event of organ dysfunction and finally, septic shock in case of refractory hypotension \cite{levy20032001}.

Rule-based disease severity scoring systems such as SOFA \cite{vincent1996sofa}, qSOFA \cite{singer2016third,seymour2016assessment}, MEWS \cite{subbe2001validation} etc. are often used in hospitals for quantitative definition of the onset of sepsis but these scores lack credibility in sepsis diagnosis. With the increase in publicly available Electronic Health Records (EHR), it is possible to design an efficient and robust system to not only have an accurate detection but also a reliable prediction of sepsis.

In this paper, we study the performance of various ML algorithms on the publicly available MIMIC-III dataset for detection and prediction of three categories of sepsis. We explore ML and DL architectures that perform well for sepsis detection and prediction and compare their performance against standard rule-based models. Finally, we rank features to understand how the vital signs are correlated with sepsis and what combination of vitals provides the best detection and prediction performance. Our contributions can be summarized as follows:
\begin{itemize}
    \item We provide the first benchmark of three categories of sepsis detection and prediction on the MIMIC-III dataset with different rule-based and ML methods.
    \item We demonstrate a deep learning based ensemble method that achieves the highest AUC compared to the single models and other ensemble models in most cases.
    \item Our ensemble model achieves the highest AUC for three categories of sepsis detection and prediction on MIMIC-III ICU patient dataset. We also show the highest improvement of AUC over the rule-based methods in the literature.
    \item We provide a feature ranking for identifying which features are most important for sepsis prediction. 
    \item We show that there is a universal improvement in AUC when six vital signs are used for both detection and prediction tasks. 
\end{itemize}

The paper is organized as follows: in section \ref{relWork}, we review the relevant research in the literature, in section \ref{experiments}, we describe the details of the experiment from data pre-processing to network design. In section \ref{results}, we show the main results in the paper and in section \ref{discussion}, we discuss our results in comparison to other works done in the literature. Finally in sections \ref{limitation} and \ref{conclusion} we discuss the limitations, future extensions and concluding remarks. 

\section{Related Works}
\label{relWork}

Use of EHR, laboratory results, and biomedical signals to track patients' sepsis progression from one stage to another 
to prevent fatal injury and death in the intensive care unit is a common approach \cite{ho2012imputation, henry2015targeted, thiel2010early, henry201463, shavdia2007septic, gultepe2013vital}. Some studies focused on a viable way of calculating the mortality rate of sepsis patients \cite{marty2013lactate, charles2014predicting, ford2016severe,calvert2017cost,che2018recurrent,purushotham2017benchmark,harutyunyan2017multitask} while Systemic Inflammatory Response Syndrome (SIRS) criteria  \cite{lukaszewski2008presymptomatic, bone1992definitions, jones2014spontaneous, kim2010temporal, calvert2016computational} or high frequency heart rate variability \cite{fairchild2013predictive, bravi2012monitoring} was used by others to predict sepsis by analyzing before and after-onset symptoms. 

In \cite{lukaszewski2008presymptomatic}, blood analysis by RT-PCR expression and neural network analysis of related genes were performed to predict sepsis onset for 92 ICU patients. This study managed to predict 83.09\% of cases 1 to 4 days prior to the clinical diagnosis with a sensitivity and specificity of 91.43\% and 80.20\% respectively. In another study, sepsis onset was predicted 2 to 3 days prior to the diagnosis through cell motion analysis using microfluidic devices \cite{jones2014spontaneous}. \cite{kim2010temporal} used lab tests, biomedical signals, and SIRS scores to create a support vector machine (SVM) model that predicts the onset 0-24 h before diagnosis in 1,239 postoperative patients of which only 26 patients (2.1\%) had sepsis, indicating a huge data imbalance that even 100 bootstraps could not address properly. The AUC ranged between 0.28 and 0.95 and the authors didn't report the accuracy of each group. These processes of sepsis detection i.e. using expensive medical equipment, acquiring daily blood sample, lab test results or performing gene analysis are not practical for regular usage.

\cite{calvert2016computational} proposed a machine learning model with gradient tree boosting for 3 hours early prediction of sepsis, called InSight. This model takes nine items extracted from patient information, laboratory test results, and widely used vitals. They used 1,394 patients from a medical intensive care unit (MICU) of which 159 patients had sepsis. Their reported sensitivity, specificity, and AUC are 0.90, 0.81, and 0.83 respectively. \cite{calvert2017cost} and \cite{desautels2016prediction}  also used the same Insight model for severe sepsis detection and got an AUC of 0.89 at the onset and 0.75, 4 hours prior to the onset. All these works either used MIMIC-III or MIMIC-II dataset. \cite{mao2018multicentre} validated the InSight primarily on a mixed-ward retrospective dataset from the University of California, San Francisco (UCSF) Medical Center (San Francisco, CA) for detection and prediction of three sepsis related gold standards and got an AUC of 0.92 and 0.87  for detection of sepsis and severe sepsis respectively. This time InSight used six clinical items. \cite{kam2017learning} used deep learning models to make early sepsis prediction system and verification of its feature extraction capacity. The best result they got was an AUC of 0.929, using a variant of LSTM. They followed the feature extraction steps of \cite{calvert2016computational}.     

\section{Experiments}
\label{experiments}

\subsection{Dataset}

This work uses the Medical Information Mart for Intensive Care (MIMIC)-III version 1.4 dataset \cite{johnson2016mimic}, compiled from the patients admitted to the Beth Israel Deaconess Medical Center (BIDMC) in Boston, MA between 2001 and 2012. The MIMIC-III set includes comprehensive clinical data such as vital signs, medications, laboratory measurements, fluid balance, procedure codes, diagnostic codes, imaging reports, hospital length of stay, survival data etc. from over 53,423 ICU stays for more than 40,000 patients. Two different critical care information systems CareVue (Philips) and Metavision (iMDSoft) were used for data entry, which handle and store some information differently. These systems were in place from 2001 to 2008 and 2008 to 2012, respectively. We used only the EHR-entered components of the MIMIC-III dataset, without any real-time waveform data or free text notes. Since the original MIMIC-III data collection did not impact clinical care and all data were deidentified in accordance with the Health Insurance Portability and Accountability Act (HIPAA) standards, the requirement for individual patient consent was waived by the Institutional Review Boards of BIDMC and the Massachusetts Institute of Technology.

\subsection{Data Extraction and Imputation}

The data were provided in the form of comma separated value (CSV) files and stored in a PostgreSQL \cite{PostgreS89:online} database. All the necessary CSV files were downloaded following the instructions mentioned in \cite{MIMICIII52:online}. Several python \cite{van2003python} scripts were written to extract measurements and patient outcomes of interest cross-matching relevant CSV files. For each patient, all the measurements were binned by hour. For patients without any measurement in a given hour, the missing measurement was filled in using carry-forward imputation, that is applying the patient’s last measured value to the following hour. When the patient didn't have any measurement prior to the missing hour, it was filled with the next available measurement. In the case of multiple measurements within an hour, the mean was used. After the data were processed, they were used to train, test and compare several machine learning classifiers to predict sepsis at the positive hours and 4 hours prior to that.

\subsection{Gold Standards}
\label{gstandards}

For sepsis, severe sepsis and septic shock, we followed the gold standard and onset time definitions as mentioned in \cite{mao2018multicentre} to create our sepsis, severe sepsis and septic shock dataset for training and testing purposes. In fact, we were more concerned about the whole time series data of each patient's entry. So in stead of stopping just at the onset, we looked for all hours where vitals met the same standards. We call these the positive hours. Some conditions such as organ dysfunction, chronic dialysis, pneumonia, and kidney injury were ignored for severe sepsis to avoid complexity.

\subsection{Comparators}
\label{comp}

We compared our best model's performance for each gold standard to three common severity scoring systems: SOFA, qSOFA, and MEWS. To calculate the SOFA score, we took each patient’s PaO2/FiO2, Glasgow Coma Score (GCS), mean arterial blood pressure, dobutamine, epinephrine, norepinephrine, dopamine dosage, bilirubin level, platelet counts, and creatinine level from five different CSV files. Then we categorized them in 6 categories and scored from 1 to 6 as described in \cite{vincent1996sofa}. The total sum of the category scores may go up to 24. For qSOFA, we used GCS, respiratory rate and systolic blood pressure as mentioned in \cite{singer2016third}. The score ranges from 0 to 3. Finally, the MEWS score, which ranges from 0 (normal) to 14 (high risk), was determined from heart rate, systolic blood pressure, respiratory rate, temperature, and GCS. We used the scoring system presented in \cite{lam2006validation} to compute each patient’s MEWS score.

\subsection{Inclusion Criteria}

For this study, we considered six clinical vital sign measurements: heart rate, peripheral capillary oxygen saturation (SpO2), respiratory rate, systolic blood pressure, diastolic blood pressure, and temperature. We used only vital signs, which are frequently available and routinely taken in the ICU, ED, and floor units. Patient data were used from the course of a patient’s hospital encounter. Patients in our final datasets were required to 
\begin{itemize}
    \item be adult (i.e. age $\geq$ 18).
    \item be admitted to medical Intensive Care Unit (ICU).
    \item have at least one measurement for each of the six vital signs.
    \item have at least 7 hours of data before the onset.
\end{itemize}  
Patients who didn't meet any of the above criteria, were excluded. Multiple hospital admissions (hadms) of the same patient were considered as separate entries. 

After meeting all the above requirements and gold standards as mentioned in \ref{gstandards}, the final dataset contained 299 entries (out of 1240 hadms from 288 patients) for sepsis, 1046 entries (out of 3788 hadms from 1012 patients) for severe sepsis and 493 entries (out of 2520 hadms from 485 patients) for septic shock. Also we randomly chose 1000 entries from non-sepsis patient hadms for the final dataset.

\subsection{Feature Selection}

We took six raw vital sign data to generate our features. Following all the above-mentioned steps, we obtained three hourly values for each of the six vital sign measurements from the positive hour, the hour prior, and two hours prior. Then we took the two differences for these three measurements. That made a total of five values from each vital sign resulting in a final feature vector of 30 elements for each positive hour in an entry of our final dataset. This feature generation process was adopted from \cite{mao2018multicentre}. The difference values i.e. gradient information helped our classifiers to capture the temporal nature of the data. 

\subsection{Model Design}

We performed two tasks - detection and prediction. Detection was done on the positive hours and prediction was done 4 hours prior to the positive hours. Positive hour is defined in section \ref{gstandards}. With our intuitive understanding of the data structures, we explored 4 classifiers for performance comparisons. They are Logistic Regression (LR), Random Forest (RF), XGBoost (XGB), and feedforward neural network or Multilayer Perceptron (MLP). 

Logistic regression and XGBoost were implemented in python, using the scikit-learn  \cite{scikitle53:online} and XGBoost \cite{PythonPa50:online} python packages respectively. Random Forest was implemented in Weka version 3.8 \cite{Weka3Dat38:online}. We tuned the hyper-parameters to obtain the most optimized models. We designed our neural networks in python. The chosen framework was keras \cite{HomeKera33:online} with tensorflow \cite{TensorFl71:online} at back-end. Keeping the sample size and input feature dimension in mind, we designed a shallow 3 layer neural network. Due to the very small size of our datasets, more layers are redundant as that would eventually make the model overfit within very few epochs. The output value of our MLP denoted as $o$, can be expressed as,

\begin{align} 
o = f( W_{out}^{T}f(W_2^{T}f( W_1^{T}x+b_1)+b_2)+b_{out}) 
\end{align}
where $x$ is the input feature vector, $W_1$, $W_2$, $W_{out}$ are the weight matrices and $b_1$, $b_2$, $b_{out}$ are the bias terms of the 1st, 2nd and output layers respectively. The number of neurons on each layer, learning rate, optimizer, activation functions, regularization coefficient, and dropout rate were tuned from multiple runs for each of the six cases (two tasks for each of the three categories). We found different sets of hyper-parameters working better for different cases. There was no universal set to work better in all cases.

Finally, we explored the ensemble of models using the best three models that are Random forest, XGBoost, and MLP. There are a number of ways to perform ensemble on the trained models including linear averaging, bagging, boosting, stacked regression etc. In our previous work, we obtained the best results using simple linear averaging of the probabilities given by the individual models \cite{islam2017abnormality}. In this work, we also explored a new architecture with a second 3 layer MLP that we trained on the 3 model's binary output probabilities. So this network takes a 6-dimensional probability vector as input rather than the 30-dimensional feature vector like any other single model. This model performed the best compared to all the other classifier models.

For all cases, we split our datasets in 70:10:20 ratio, that is we considered 70\% of the data for training, 10\% for validation to tune model hyper-parameters and rest 20\% for testing. 

\section{Results}
\label{results}

We will report our results in two steps. First, we will compare all the classifiers' performance and then we will compare the best model for each case with three standard scoring systems - SOFA, qSOFA and MEWS. 

\subsection{Evaluation Metrics}

The performance of our classifier models was evaluated in terms of area under receiver operating characteristics curve (AUC). ROC curve is the graphical plot of true positive rate (TPR) vs false positive rate (FPR) of a binary classifier when classifier threshold is varied from $0$ to $1$. The number of positive instances that are correctly identified by the classifier is called true positive (TP) and the number of positive instances that are incorrectly classified by the classifier is called false negative (FN). The number of normal instances that are correctly classified as normal is called true negative (TN), and in a similar fashion, the number of normal instances that are incorrectly identified as positive instances is called false positive (FP). True positive rate (TPR) or sensitivity is the proportion of positive instances that are correctly identified as positive instances, while false positive rate (FPR) is the proportion of normal instances that are incorrectly identified as positive instances. 

TPR or sensitivity shows the degree to which a model does not miss a positive instance. On the other hand, specificity indicates the degree to which a model correctly identifies normal instances as normal. The objective of a model is to attain high sensitivity as well as specificity so that it attains low diagnosis error. 

\subsection{Classifiers comparisons}

The results have been reported in Table \ref{CompRp}. As we can see, from the results, out of the 4 base classifiers, RF is the clear winner in most of the cases while LR did the worst. XGB and MLP lie in between. We believe it's lack of enough data that caused the MLP to perform weakly compared to RF. But when we tested our ensemble model, it surpassed the best scores of the 4 classifiers. In detection task, it achieved an AUC of 0.97, 0.96 and 0.91 for sepsis, severe sepsis and septic shock respectively. 
Out of the three categories using our ensemble model, we got a 1\% increase in both sepsis and septic shock compared to the best classifier performance. In prediction task, our ensemble model got an AUC of 0.90, 0.91 and 0.90 for the three categories respectively which are again better (2\%, 1\% and 1\% increase) than the best classifier scores. 

\begin{table}[!h]
\caption{AUC using different classifiers for three sepsis gold standards. Here, Det = Detection, Pred = Prediction, S = Sepsis, SS = Severe Sepsis, SK = Septic Shock, MLP = Multilayer Perceptron, LR = Logistic Regression, XGB = XGBoost, RF = Random Forest.}
\label{CompRp}
\vskip 0.15in
\begin{center}
\begin{small}
\begin{tabular}{|c|c|c|c|c|c|c|}
\hline
Task & Category & MLP & LR & XGB & RF & Ensemble\\
\hline
\multirow{3}{*}{Det}
& S   & 0.91 & 0.86 & 0.96 & 0.96 & \textbf{0.97}\\
& SS  & 0.92 & 0.85 & 0.96 & 0.96 & \textbf{0.96}\\
& SK  & 0.86 & 0.75 & 0.90 & 0.90 & \textbf{0.91}\\ \hline
\multirow{3}{*}{Pred}
& S   & 0.84 & 0.76 & 0.88 & 0.88 & \textbf{0.90}\\
& SS  & 0.86 & 0.76 & 0.90 & 0.90 & \textbf{0.91}\\
& SK  & 0.85 & 0.73 & 0.89 & 0.89 & \textbf{0.90}\\
\hline
\end{tabular}
\end{small}
\end{center}
\vskip -0.1in
\end{table}

We can notice two interesting trends here. Unlike other two categories, for septic shock there is almost no significant difference in detection and prediction tasks. This might be attributed to the fact that, septic shock being the ultimate stage of septicemia, has more informative vitals for all the classifiers, even 4 hours prior to the positive hours, compared to the other two categories. For example a patient who has sepsis now (detection task), might not develop any relevant symptom for sepsis 4 hours ago (prediction task). But that's not the case for septic shock, as a patient diagnosed with septic shock now has a high chance of developing sepsis or severe sepsis 4 hours ago. Another point worth mentioning is the very little improvement in severe sepsis. We believe this is due to not having the necessary organ failure measurements as inputs. More about this in section \ref{discussion}.

\subsection{Comparison with Benchmarks}

Next, we compared our ensemble model with three standard severity scoring systems - SOFA, qSOFA, and MEWS. We calculated AUC 
for all 3 systems following the procedures mentioned in section \ref{comp}. The results have been presented in Table \ref{FinalCompRp}. Out of the three systems, MEWS did the best in detection task (0.72, 0.76 and 0.66 for sepsis, severe sepsis, and septic shock respectively) and SOFA did the best in prediction task (0.54, 0.60 and 0.64 for sepsis, severe sepsis and septic shock respectively). It's clear that none of the scoring systems are reliable enough to detect sepsis, severe sepsis or septic shock let alone predict. In fact, even our ensemble model's prediction scores are significantly higher than the best detection scores of the three scoring systems and this holds true for all three cases. 

\begin{table}[!h]
\caption{Comparison with rule-based scoring systems in term of AUC. Here, Det = Detection, Pred = Prediction, S = Sepsis, SS = Severe Sepsis, SK = Septic Shock.}
\label{FinalCompRp}
\vskip 0.15in
\begin{center}
\begin{small}
\begin{tabular}{|c|c|c|c|c|c|c|}
\hline

Task & Category & SOFA & qSOFA & MEWS & Ensemble\\
\hline
\multirow{3}{*}{Det}
& S   & 0.62 & 0.66 & 0.72 & \textbf{0.97}\\
& SS  & 0.66 & 0.72 & 0.76 & \textbf{0.96}\\
& SK  & 0.63 & 0.61 & 0.66 & \textbf{0.91}\\ \hline
\multirow{3}{*}{Pred}
& S  & 0.54 & 0.44 & 0.49 & \textbf{0.90}\\
& SS & 0.60 & 0.56 & 0.59 & \textbf{0.91}\\
& SK & 0.64 & 0.57 & 0.63 & \textbf{0.90}\\ \hline
\end{tabular}
\end{small}
\end{center}
\vskip -0.1in
\end{table}

\section{Discussions}
\label{discussion}
There have been many works regarding early detection i.e. prediction of sepsis \cite{ho2012imputation, henry2015targeted, thiel2010early, henry201463, shavdia2007septic, gultepe2013vital,lukaszewski2008presymptomatic,jones2014spontaneous,kim2010temporal} but they used either laboratory test results or expensive equipment for data analysis which significantly increase cost and delay the whole process of detection - making the systems practically unfeasible. There are also some works \cite{marty2013lactate, charles2014predicting, ford2016severe,calvert2017cost,che2018recurrent,purushotham2017benchmark,harutyunyan2017multitask} where mortality rate was calculated for sepsis patients. Considering the choice of features or final objective, all these works are somewhat irrelevant to our study.

In \cite{calvert2016computational}, authors proposed an early sepsis detection system, called InSight and validated it on MIMIC-II dataset. 3 hours prior to the onset, they achieved an average AUC of 0.83 which is a bit higher than our AUC score for sepsis in prediction task. But we also need to account for the differences in our studies such as different dataset (MIMIC-II vs MIMIC-III), different data preprocessing schemes and feature selection process and most importantly different hour look-ahead (3 vs 4). Motivated by \cite{calvert2016computational}, \cite{kam2017learning} experimented the effects of MLP and LSTM models on prediction tasks following the same data processing pipeline. They used three different feature vector sets of dimension 100, 109 and 209 which are a lot higher compared to our 30-dimensional feature vector for each patient. Both of these works reported results only for sepsis with a different gold standard definition, contrary to our results for all 3 categories.

\cite{desautels2016prediction} also used InSight to assess its performance for severe sepsis, on the population of MIMIC-III who were logged using MetaVision. Their reported AUCs are 0.89 in detection and 0.75 in prediction task which are lower compared to our findings (0.91 in detection and 0.81 in prediction). However, their gold standard definition, patient inclusion procedures, and feature vectors are significantly different than ours.

Most relevant to our work is the study presented in \cite{mao2018multicentre}. Our selection of features and gold standard definitions were also inspired by this study. The authors reported sepsis detection scores for the three categories of sepsis and prediction score for severe sepsis only. However, their performance scores were reported on private datasets and hence, cannot be compared directly. Our work advances the state of the art in two ways compared to \cite{mao2018multicentre}. We provide comprehensive benchmark performance of various rule-based and ML models on the publicly available MIMIC-III dataset and demonstrate an ensemble model that performs better than any other single or ensemble model. Second, we do feature ranking to show that different vital sign signal is ranked higher for different single or ensemble models. However, there is a universal performance improvement for all single and ensemble models when the number of vital signs is increased from 5 to 6. We discuss these two observations in the two subsections below. 

\subsection{Benchmark for Three Categories of Sepsis}

We report the first benchmark of three different ML methods i.e. LR, XGB and RF and three rule-based methods i.e. SOFA, qSOFA, MEWS on all three categories of sepsis for detection and prediction task. Availability of these benchmark numbers on a publicly available dataset enables future works to be compared against. 

In addition, we applied neural network models for the first time on this task. We found other tasks i.e. mortality prediction or severity scoring from early admission data for which neural network almost always outperform logistic regression \cite{clermont2001predicting,jpm2040138,NIPS1995_1081}. \cite{purushotham2017benchmark} and \cite{harutyunyan2017multitask} worked extensively with deep neural networks on MIMIC-III for different prediction tasks such as in-hospital mortality, length of stay, ICD-9 code group etc. In particular, \cite{purushotham2017benchmark} used different sets of features and different types of algorithms including deep learning models to show the effectiveness of deep learning on such datasets. The authors showed that for multiple data modalities, specially when a large number of raw clinical time series data is used as input features, deep models learn better feature representations and this held true for all three tasks they performed. Both these works demonstrate the performance benefit of deep learning compared to other ML methods on this dataset. However, in our study, single neural network architectures performed poorly compared to RF. We varied the number of hidden units and layer numbers to make sure that the model has enough capacity to learn from the data features while avoiding over-fitting. However, for all the single NN architecture and sizes that we explored, we found RF to have higher AUC than NN in most of the cases. 

In order to boost AUC, we explored standard ensemble techniques - averaging and weighting. None of them showed any significant improvement. So we designed a second 3 layer MLP that we trained on the 3 model's binary output probabilities. This ensemble model design performed better than any single model and provided higher AUC consistently for all the tasks studied in this work. For example averaging gave us an AUC of 0.96, 0.96, 0.90 for 3 categories of sepsis in detection task which is basically no improvement over the best model's result. Here our ensemble model achieved improvement of 1 percentage point for both sepsis and septic shock compared to the averaging. Similarly, we got 2, 1 and 1 percentage improvements respectiely in prediction task for all three categories compared to the averaging. Our ensemble model outperformed standard disease severity scores such as SOFA, qSOFA, and MEWS for both the detection and prediction of sepsis, severe sepsis, and septic shock. This is one of the main contributions of this paper. To the author's knowledge, this is the first study that incorporates deep learning and ensemble design for the detection and prediction of all three sepsis categories, taking only six vital measurements as inputs. 

A direct comparison of the performance of our ensemble model can't be performed with the results presented in \cite{mao2018multicentre} as those results are reported on private data collected from UCSF. One comment can be made about the overall improvement of AUC by using ML model compared to the rule-based model. Whereas \cite{mao2018multicentre} achieved AUC improvement of 5 to 11 percentage points compared to the rule-based models, our ensemble model achieves an improvement of 20 to 36 percentage points by using ensemble models compared to the rule-based models. This large margin of improvement is a strong motivator for using this type of ensemble models for sepsis detection and prediction in ICU patients. 

\subsection{Feature Ranking}

We ranked the six vital sign input stream to gauge their individual effectiveness in sepsis detection and prediction. The results are shown in Table \ref{FRank}. We numbered the six vital measurements 1-6 in the following order - heart rate, spO2, respiratory rate, systolic blood pressure, diastolic blood pressure and temperature. Here we reported result for two cases- detection of sepsis and prediction of septic shock. We found feature 4 (systolic blood pressure) and feature 6 (temperature) as the most important vital signs for prediction and detection respectively. In \cite{mao2018multicentre}, the authors also found systolic blood pressure as the most important vital sign on MIMIC-III for similar tasks. Also a close observation reveals, feature 3 (respiratory rate) and feature 1 (heart rate) as the next most important features.

\begin{table}[!h]
\caption{Features ranking in term of AUC. Systolic blood pressure and temperature are the most important vital signs for sepsis prediction and detection respectively.}
\label{FRank}
\vskip 0.15in
\begin{center}
\begin{small}
\begin{tabular}{|c|c|c|c|c|c|c|}
\hline

Feature No. & Sepsis Detection & Septic Shock Prediction\\
\hline
1 & 0.85 & 0.75 \\
2 & 0.82 & 0.70 \\
3 & 0.87 & 0.74 \\
4 & 0.80 & \textbf{0.79} \\
5 & 0.81 & 0.69 \\
6 & \textbf{0.90} & 0.71\\ \hline
\end{tabular}
\end{small}
\end{center}
\vskip -0.1in
\end{table}

\begin{table}[!h]
\caption{Features ablation in term of AUC. Sepsis is highly correlated with six vital signs. All models perform significantly better when six vital signs are used.}
\label{FAb}
\vskip 0.15in
\begin{center}
\begin{small}
\begin{tabular}{|c|c|c|c|c|c|c|}
\hline

\# of Features & Sepsis Detection & Septic Shock Prediction\\
\hline

1 & 0.91 & 0.79\\
2 & 0.90 & 0.80\\
3 & 0.91 & 0.80\\
4 & 0.92 & 0.83\\
5 & 0.92 & 0.83\\
6 & \textbf{0.97} & \textbf{0.90}\\ \hline

\end{tabular}
\end{small}
\end{center}
\vskip -0.1in
\end{table}

We also increased the number of vital signs one by one to see if there is a specific trend in AUC improvement. In Table \ref{FAb}, the number of vitals for set 2-6 has been chosen in a way so that the gradual change, upon addition of new vital, becomes clear. We find that there is no particular trend up to five vital signs. We have performed different combinations of vital signs to ensure that there is indeed no specific trend of AUC improvement when five vital signs are used as input. However, we find that there is a universal improvement in AUC when six vital signs are used for both detection and prediction tasks. It is not clear exactly why six vital signs provide a universal jump in AUC for all the models be it single or ensemble. This is a topic of future exploration.

\section{Limitations and Future Scopes}
\label{limitation}
In this section, we remark few shortcomings of the present study and the possibility of future improvements. Most of the shortcoming stem from the collection and quality of ICU data. 
\begin{itemize}
    \item The MIMIC-III dataset was derived from only one institution. So it's not possible to claim universal adaptability of our models to other populations on the basis of this study alone. However, since neural networks learn from the data, similar architectures should perform well when trained on data from different demographics. 
    \item Our gold standard references to determine sepsis, severe sepsis, and septic shock rely on ICD-9 codes which might fail to capture all septic patients in the dataset if there were undetected sepsis patient. This is a limitation of the process itself, though ICD-9 codes have been used before for accuracy validation in the detection of severe sepsis \cite{Iwashyna2014IdentifyingPW}. 

    \item The sequence of laboratory tests mostly depends on physician suspicion. As a consequence, the gold standards are highly subjective and dependent on individual physician. A consistent definition of proper gold standard generation is a task for future.

    \item Our imputation process and averaging of all measurements in an hour's interval may lead to the loss of some temporal information which might affect the performance of our models. For time-series data like this study, \cite{che2018recurrent} proposed a better imputation mechanism that can capture missing information. Performance of our model on a better imputed data stream is a topic of future exploration. 

    \item Our trained models require at least three hours of data to predict or detect; thus eliminating the possibility of any first or second hour evaluation of any patient. We intend to work on these in our future work.
\end{itemize}

\section{Conclusion}
\label{conclusion}
In this study, we have provided the first benchmark of various ML and rule-based models for three categories of sepsis detection and prediction task on the MIMIC-III dataset ICU patient population. We then designed a particular ensemble model that outperformed all the single model results. Our ensemble model also showed a large margin of improvement over common rule-based sepsis detection methods. We rank the vital signs to find that six vital signs provide better performance than any other combination of vital signs for all the models. Since the model uses only six vital signs, we believe our model will be useful for application in real-world hospital environment.

\section{Acknowledgement}
\label{acknowledge}
This research used resources of the National Energy Research Scientific Computing Center, a DOE Office of Science User Facility supported by the Office of Science of the U.S. Department of Energy under Contract No. DE-AC02-05CH11231. Thanks Leonid Oliker at NERSC for sharing his allocation on the OLCF Titan supercomputer with us on project CSC103. Thanks to Prabhat at NERSC for sharing his allocation on the NERSC computers. Thanks to Uli Chettipally, MD for some initial discussions regarding sepsis detection. 
\bibliography{refs}
\bibliographystyle{IEEEbib}

\clearpage

\end{document}